\documentclass{files/ieeeconf}
\usepackage{cite}
\usepackage{amsmath,amssymb,amsfonts}
\usepackage{graphicx}
\usepackage{xcolor}
\usepackage[acronym]{glossaries} 

\usepackage{files/customCommands} 
\usepackage{files/paperCommands} 

\def\thesubsubsectiondis{\unskip\arabic{subsubsection})}

\def\BibTeX{{\rm B\kern-.05em{\sc i\kern-.025em b}\kern-.08em
    T\kern-.1667em\lower.7ex\hbox{E}\kern-.125emX}}

\begin{document}

\title{Borinot: an agile torque-controlled robot\\ for hybrid flying and contact loco-manipulation}

\author{Josep Mart\'i-Saumell \quad Joan Sol\`a \quad Angel Santamaria-Navarro \quad Hugo Duarte 
\thanks{All authors are with the Institut de Rob\`otica i Inform\`atica Industrial, CSIC-UPC, Llorens Artigas 4-6, Barcelona 08028 (e-mail: \{nsurname\}@iri.upc.edu). This work was partially supported by the Spanish Ministry of Science and Innovation under the projects EBCON (PID2020-119244GB-I00, funded by MCIN/ AEI /10.13039/501100011033 ); AUDEL (TED2021-131759A-I00, funded by MCIN/ AEI /10.13039/501100011033 and by the "European Union NextGenerationEU/PRTR") and by the Consolidated Research Group RAIG (2021 SGR 00510) of the Departament de Recerca i Universitats de la Generalitat de Catalunya.}
}
\maketitle


\begin{abstract}

This paper introduces Borinot, an open-source flying robotic platform designed to perform hybrid agile locomotion and manipulation.
This platform features a compact and powerful hexarotor that can be outfitted with torque-actuated extremities of diverse architecture, allowing for whole-body dynamic control. 
As a result, Borinot can perform agile tasks such as aggressive or acrobatic maneuvers with the participation of the whole-body dynamics.

The extremities attached to Borinot can be utilized in various ways; during contact, they can be used as legs to create contact-based locomotion, or as arms to manipulate objects. In free flight, they can be used as tails to contribute to dynamics, mimicking the movements of many animals. This allows for any hybridization of these dynamic modes, like the jump-flight of chicken and locusts, making Borinot an ideal open-source platform for research on hybrid aerial-contact agile motion.

To demonstrate the key capabilities of Borinot, we have fitted a planar 2DoF arm and implemented whole-body torque-level model-predictive-control. The result is a capable and adaptable platform that, we believe, opens up new avenues of research in the field of agile robotics.
\end{abstract}

\section{The agile robot}

The field of robotics is experiencing a surge in interest in the comprehension and creation of complex and dynamic motion~\cite{Rubenson_SR2022, Ajanic_SR2020, foehn_AgiliciousOpensourceOpenhardware_2022}. This understanding is not only fascinating from a biomechanical standpoint, but also essential in effectively building and controlling naturally unstable robots like humanoids and aerial robots. This is especially evident in the realm of humanoid and legged robotics, where the multi-articulated body parts interact dynamically to generate overall motion.
The rise of unmanned aerial manipulators (UAMs) has resulted in aerial robots becoming increasingly multi-articulated~\cite{Ollero_2022}. This presents new possibilities, such as aerial loco-manipulation using contacts, the action of tail mass to modify flight dynamics, or some hybrid locomotion modes such as the jump-and-fly technique employed by certain animals like chickens and locusts.

In this communication we introduce Borinot, an agile aerial loco-manipulator designed to conduct research on high dynamic motion. We shall first discuss our notion of agility, departing from some initial definition and refining it to be able to engineer it. We will discuss how such concept of agility can be implemented in robots, both in terms of electro-mechanics and control. We will finally demonstrate some of the abilities of Borinot through the real execution of diverse motion modes.

\begin{figure}[t]
\centering
\includegraphics[width=0.7\columnwidth]{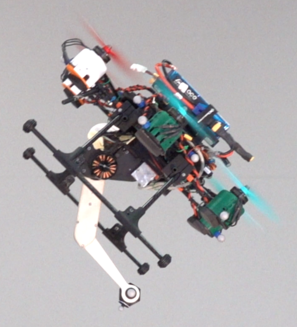}
\caption{The open-source Borinot platform for agile whole-body torque-controlled flying and contact locomotion.}
\label{fig:borinot}
\end{figure}

\subsection{Embodiment, dynamics, and agility}

The dictionary definition of agility as the ``\textit{ability to move quickly and easily}''~\cite{agile-oxford} overlooks crucial factors for robotics. The existence of a complex multi-articulated body with 
head and limbs is taken for granted in natural language and biomechanical studies when referring to agility, but requires careful attention when engineering machines. Two characteristics of such bodies that must be considered are under-actuation and redundancy. 
Under-actuation means the state is not locally  controllable, requiring the generation of maneuvers to reach desired configurations~\cite{underactuated}. 
Redundancy means there are multiple ways to satisfy a task and thus that certain degrees of freedom remain available for secondary tasks. 
With these considerations, we shall enrich the definition of agility as ``\textit{the ability of a complex body to combine maneuvers quickly and easily}''. 

\subsection{Robot design}

This new definition outlines the necessary conditions for designing an agile robot, with each keyword representing a specific characteristic. For instance, the term `maneuver' demands strong prediction capabilities, while `quick' emphasizes dynamism. `Easy' necessitates the minimization of effort, hence the importance of optimality. Achieving quick movements with minimal effort requires the utilization of 
carefully predicted forces to modify the robot's whole-body dynamics as little as necessary. Whole-body control is also demanded by the keyword `combine'. 
To ensure easy dynamics, robots with low moment of inertia are preferred, with actuators placed where their effect is maximum.

\begin{figure}[t]
  \centering
  \includegraphics[width=0.7\linewidth]{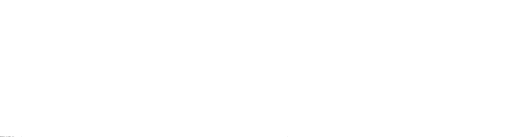}
  \vspace{-1mm}
  \caption{The Borinot control architecture combines optimal control planning, model-predictive control, and classical proportional-derivative control}
  \label{fig:exp_control_architecture}
\end{figure}

With these guidelines, Borinot (\figRef{fig:borinot}) is conceived as a powerful (5.2 thrust-to-weight ratio) and compact (370\,mm diameter) thrust-controlled hexarotor to which light and compliant torque-driven extremities \cite{grimminger_2020} can be outfitted. The center of mass is located high so as to allow rapid and easy destabilization from the hovering stance. It has a powerful control unit constituted by an Intel NUC i7 (6296 mark at cpubenchmark \cite{cpu-benchmark}), able to perform high-demanding control tasks such as whole-body force-torque-based model-predictive control (MPC, \figRef{fig:exp_control_architecture}). 

%

\section{Experiments}

\begin{figure}[t]
  \centering
  \includegraphics[width=0.8\columnwidth]{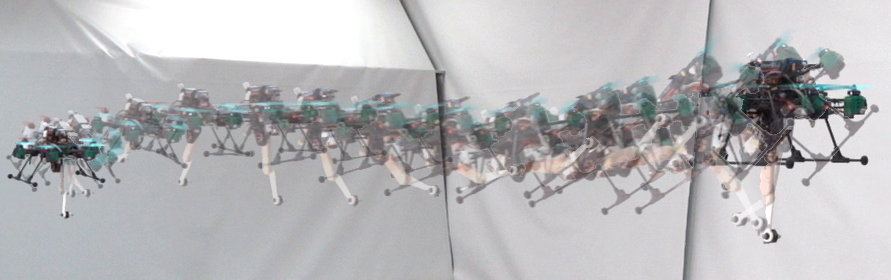}
  \centering\includegraphics[width=0.7\columnwidth]{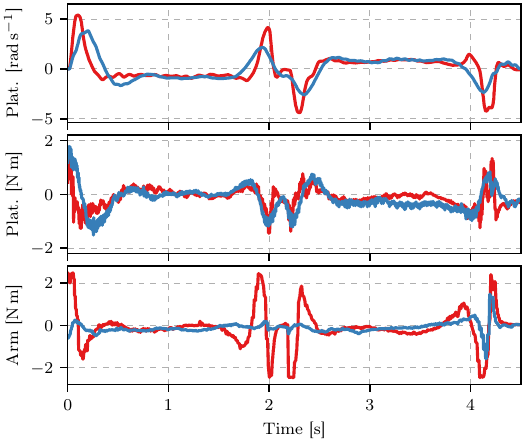}
  \vspace{-3mm}
  \caption{Tail's influence on the overall motion. Comparison between the X (red) and Y (blue) trajectories. Top: Sequence of the X trajectory. Top plot: platform's angular speed. Mid: torque due to propellers. Bottom: torque of the first joint of the tail. Notice the contribution of the tail torque on the tilting speed of the X trajectory.}
  \label{fig:exp_trajectories}
\end{figure}

We illustrate the capabilities of Borinot fitted with a planar 2DoF extremity in three agile locomotion and manipulation tasks using optimal control planning and MPC as in \figRef{fig:exp_control_architecture}.

\subsection{Flying locomotion: extremity as a tail}
\label{subsec:exp_displacement}

This task illustrates the participation of the extremity as a tail to contribute to overall dynamics (\figRef{fig:exp_trajectories}). We attached a 100\,g mass at the end effector to provide tail inertia. We ask Borinot to quickly go to a waypoint 5\,m away and back, with a short hovering pause at the waypoint. We perform two experiments: one moves along Borinot's X direction where the tail's DoF align to produce a pitching torque to the base; the other one moves in the Y direction where the tail cannot participate. In the X trajectory the platform tilts much quicker due to the dynamic assistance of the tail.

\subsection{Flying loco-manipulation: extremity as a hand}
\label{subsec:exp_end_effector}

\begin{figure}[t]
  \centering
  \scalebox{-1}[1]{\includegraphics[width=0.5\columnwidth]{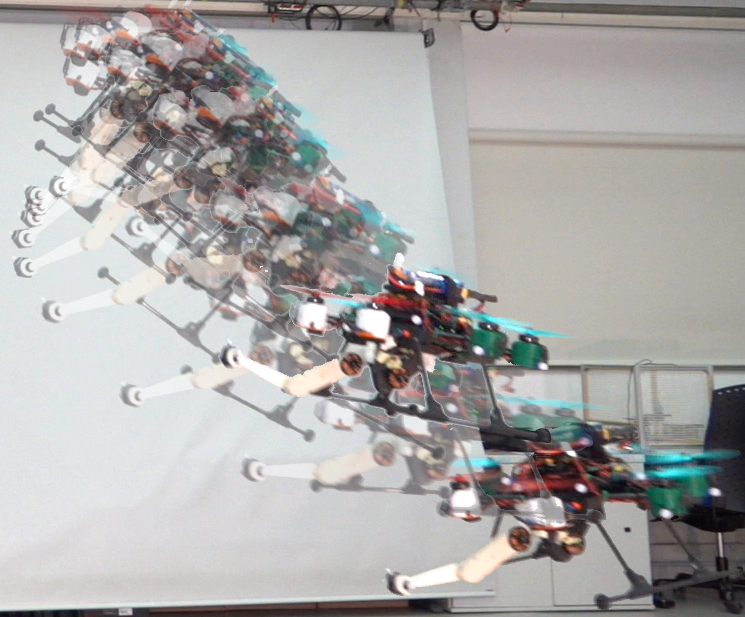}}
  \includegraphics[width=0.65\columnwidth]{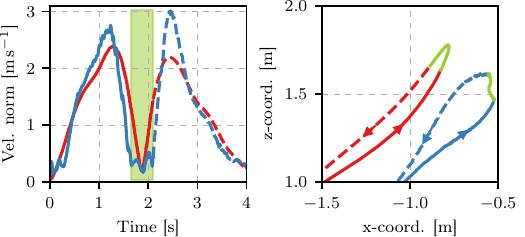}
  \vspace{-3mm}
  \caption{Image sequence of an agile positioning of the end-effector at a fixed point during unstable flight. Left: velocity profiles of base (red) and end-effector (blue) during positioning task (green). Right: XZ plane trajectories.}
  \label{fig:exp_end_effector}
\end{figure}

We illustrate in \figRef{fig:exp_end_effector} the ability of Borinot to perform a loco-manipulation task (keep the hand at a fixed point for 600\,ms) while in unstable flight (keep the base tilted $60^\circ$). Borinot is able to nearly stall the hand in place even when the base is experiencing a parabolic near free-fall trajectory.

\subsection{Jump-and-fly locomotion: extremity as a leg}
\label{subsec:jump_fly}

\begin{figure}[t]
  \centering
  \includegraphics[trim=0 0 0 22mm , clip, width=0.21\columnwidth]{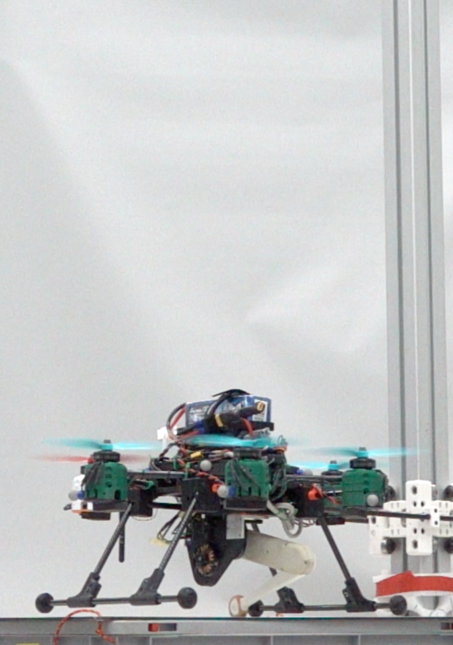}
  \includegraphics[trim=0 0 0 22mm , clip, width=0.21\columnwidth]{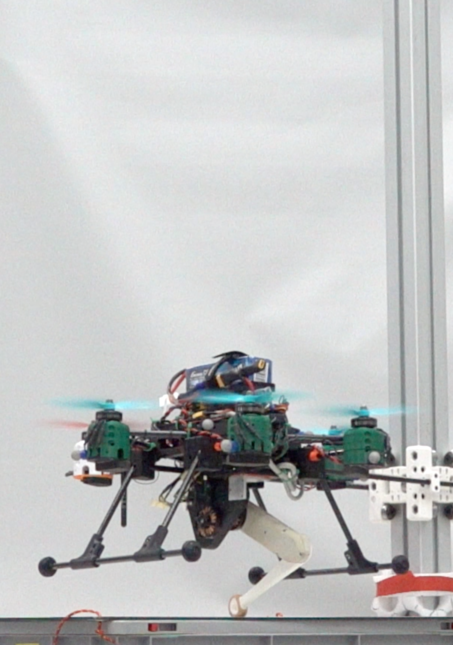}
  \includegraphics[trim=0 0 0 22mm , clip, width=0.21\columnwidth]{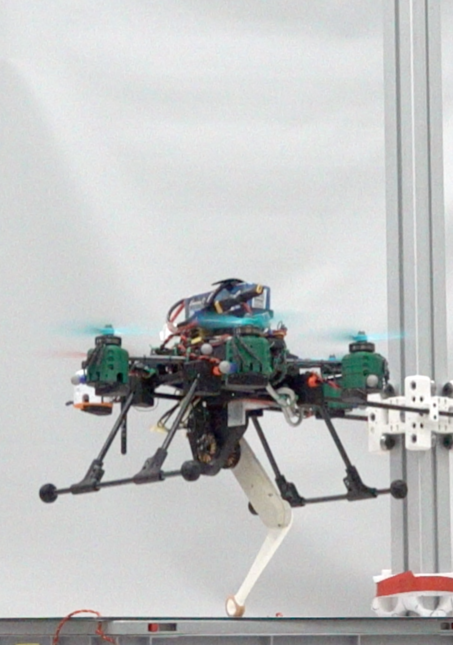}
  \includegraphics[trim=0 0 0 22mm , clip, width=0.21\columnwidth]{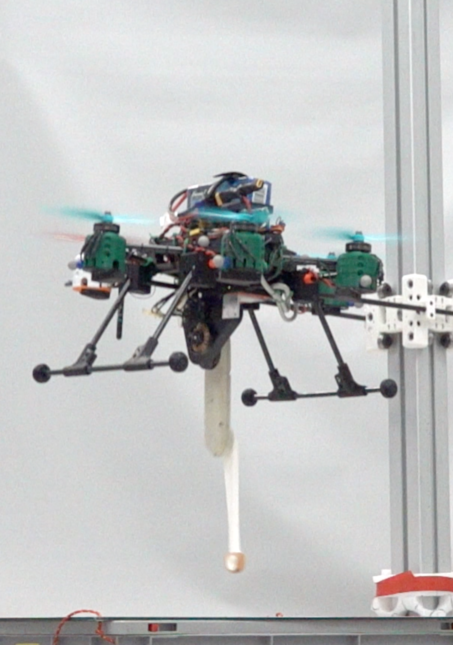}  
  \\\vspace{1mm}
  \includegraphics[trim=0 0 0 12mm , clip, width=0.21\columnwidth]{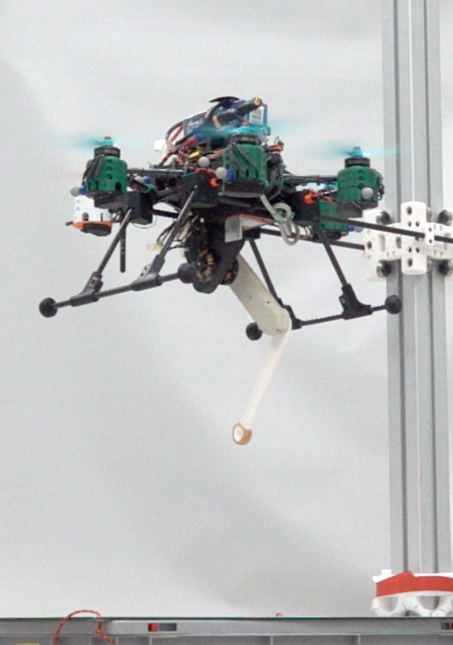}
  \includegraphics[trim=0 0 0 12mm , clip, width=0.21\columnwidth]{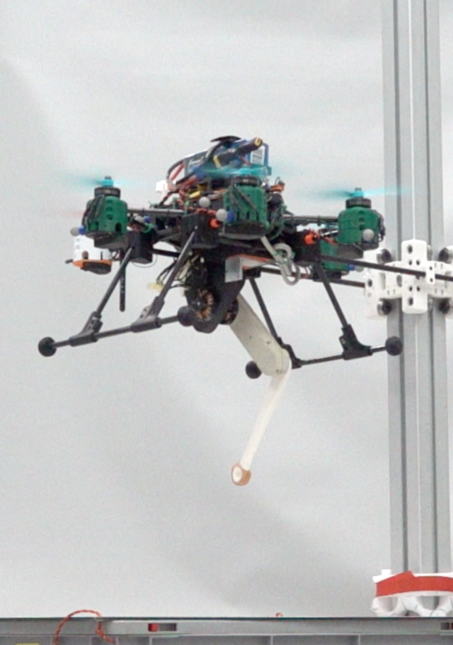}
  \includegraphics[trim=0 0 0 12mm , clip, width=0.21\columnwidth]{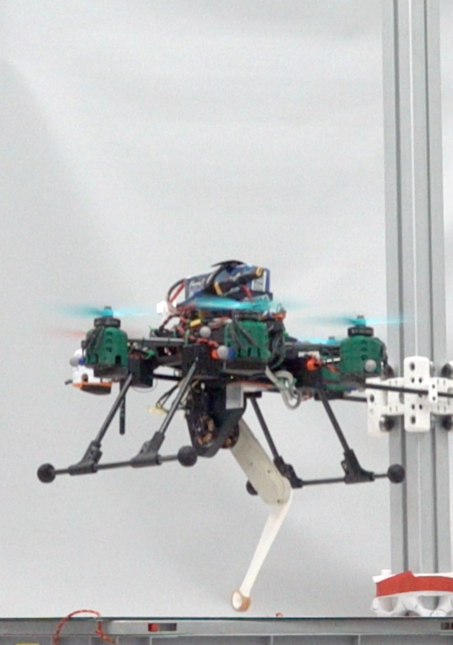}
  \includegraphics[trim=0 0 0 12mm , clip, width=0.21\columnwidth]{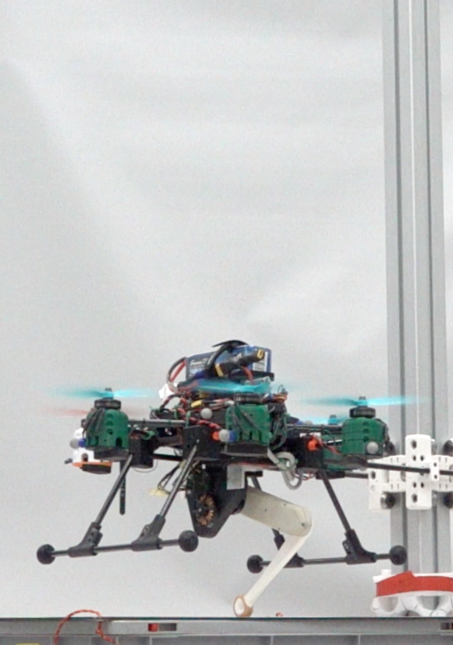}
  \caption{Jump-and-fly locomotion. Motion was constrained by a vertical guide because balance control was not implemented for this experiment. Total jump-flight time is 2.2s.}
  \label{fig:jump_fly}
\end{figure}

We provide the electro-mechanical proof for jump-and-flight motion (\figRef{fig:jump_fly}). The thrust is set to 90\% of total body weight. The leg is applied a 2.5\,Nm torque, producing a jump-flight with a slower-than-gravity parabolic fall, which is caught by the compliant leg in a smooth landing. 

\bibliography{files/references}

\begin{thebibliography}{1}
\providecommand{\url}[1]{#1}
\csname url@samestyle\endcsname
\providecommand{\newblock}{\relax}
\providecommand{\bibinfo}[2]{#2}
\providecommand{\BIBentrySTDinterwordspacing}{\spaceskip=0pt\relax}
\providecommand{\BIBentryALTinterwordstretchfactor}{4}
\providecommand{\BIBentryALTinterwordspacing}{\spaceskip=\fontdimen2\font plus
\BIBentryALTinterwordstretchfactor\fontdimen3\font minus
  \fontdimen4\font\relax}
\providecommand{\BIBforeignlanguage}[2]{{%
\expandafter\ifx\csname l@#1\endcsname\relax
\typeout{** WARNING: IEEEtran.bst: No hyphenation pattern has been}%
\typeout{** loaded for the language `#1'. Using the pattern for}%
\typeout{** the default language instead.}%
\else
\language=\csname l@#1\endcsname
\fi
#2}}
\providecommand{\BIBdecl}{\relax}
\BIBdecl

\bibitem{Rubenson_SR2022}
\BIBentryALTinterwordspacing
J.~Rubenson and G.~S. Sawicki, ``Running birds reveal secrets for legged robot
  design,'' \emph{Science Robotics}, vol.~7, no.~64, p. eabo2147, 2022.
  [Online]. Available:
  \url{https://www.science.org/doi/abs/10.1126/scirobotics.abo2147}
\BIBentrySTDinterwordspacing

\bibitem{Ajanic_SR2020}
\BIBentryALTinterwordspacing
E.~Ajanic, M.~Feroskhan, S.~Mintchev, F.~Noca, and D.~Floreano, ``Bioinspired
  wing and tail morphing extends drone flight capabilities,'' \emph{Science
  Robotics}, vol.~5, no.~47, p. eabc2897, 2020. [Online]. Available:
  \url{https://www.science.org/doi/abs/10.1126/scirobotics.abc2897}
\BIBentrySTDinterwordspacing

\bibitem{foehn_AgiliciousOpensourceOpenhardware_2022}
\BIBentryALTinterwordspacing
P.~Foehn, E.~Kaufmann, A.~Romero, R.~Penicka, S.~Sun, L.~Bauersfeld,
  T.~Laengle, G.~Cioffi, Y.~Song, A.~Loquercio, and D.~Scaramuzza,
  ``Agilicious: {{Open-source}} and open-hardware agile quadrotor for
  vision-based flight,'' vol.~7, no.~67, p. eabl6259, June 2022. [Online].
  Available: \url{https://www.science.org/doi/10.1126/scirobotics.abl6259}
\BIBentrySTDinterwordspacing

\bibitem{Ollero_2022}
A.~Ollero, M.~Tognon, A.~Suarez, D.~Lee, and A.~Franchi, ``Past, present, and
  future of aerial robotic manipulators,'' \emph{IEEE Transactions on
  Robotics}, vol.~38, no.~1, pp. 626--645, 2022.

\bibitem{agile-oxford}
\BIBentryALTinterwordspacing
``Oxford learner's dictionaries: agile,'' accessed on April 3, 2023. [Online].
  Available:
  \url{https://www.oxfordlearnersdictionaries.com/definition/english/agile}
\BIBentrySTDinterwordspacing

\bibitem{underactuated}
\BIBentryALTinterwordspacing
R.~Tedrake, \emph{Underactuated Robotics}, 2023. [Online]. Available:
  \url{https://underactuated.csail.mit.edu}
\BIBentrySTDinterwordspacing

\bibitem{grimminger_2020}
F.~{Grimminger}, A.~{Meduri}, M.~{Khadiv}, J.~{Viereck}, M.~{Wüthrich},
  M.~{Naveau}, V.~{Berenz}, S.~{Heim}, F.~{Widmaier}, T.~{Flayols}, J.~{Fiene},
  A.~{Badri-Spröwitz}, and L.~{Righetti}, ``An open torque-controlled modular
  robot architecture for legged locomotion research,'' \emph{IEEE Robotics and
  Automation Letters}, vol.~5, no.~2, pp. 3650--3657, 2020.

\bibitem{cpu-benchmark}
\BIBentryALTinterwordspacing
``Passmark software {CPU} benchmark,'' accessed on April 3, 2023. [Online].
  Available:
  \url{https://www.cpubenchmark.net/cpu\_lookup.php?cpu=Intel+Core+i7-8650U+\%40+1.90GHz\&id=3070}
\BIBentrySTDinterwordspacing

\end{thebibliography}

\end{document}